# A Temporal Neural Network Architecture for Online Learning

J. E. Smith

University of Wisconsin-Madison (Emeritus)
Carnegie Mellon University (Adjunct)
February 22, 2021

*Abstract* — **A long-standing proposition is that by emulating the operation of the brain's neocortex, a spiking neural network (SNN) can achieve similar desirable features: flexible learning, speed, and efficiency. Temporal neural networks (TNNs) are SNNs that communicate and process information encoded as relative spike times (in contrast to spike rates). A TNN architecture is proposed, and, as a proof-of-concept, TNN operation is demonstrated within the larger context of online supervised classification. First, through unsupervised learning, a TNN partitions input patterns into clusters based on similarity. The TNN learning process adjusts synaptic weights by using only signals local to each synapse, and global clustering behavior emerges. The TNN then passes a cluster identifier to a simple online supervised decoder which finishes the classification task. Besides features of the overall architecture, several TNN components and methods are new to this work.**

**The overall research objective is a direct hardware implementation. Consequently, the architecture is described at a level analogous to the gate and register transfer levels used in conventional digital design, and processing is done at very low precision.**

## I. INTRODUCTION

In the long term, this line of research is focused on the development of an online learning architecture, implemented directly in hardware, that can interact efficiently with a dynamically changing environment in an intelligent way.

The focus of the research reported here is the development of a base technology that can serve as an important steppingstone to reaching the more ambitious long term goal. The base technology processes information encoded as pulses or spikes in time and is realized as a temporal neural network (TNN) that performs unsupervised online clustering. A decoder employing online supervised learning then maps the clusters to classes. In combination, the TNN and decoder achieve online supervised classification.

### A. Motivation

Two hypotheses motivate this research, although the focus here is primarily on the first.

*First*: If the principles that underly neural computation/cognition in the brain's neocortex can be discovered, then the same principles can be applied to the development of advanced human-engineered computing devices that have brain-like capabilities and efficiencies.

This longstanding hypothesis motivated the earliest neural network research efforts. As research advanced, however, artificial neural networks (ANNs) followed a path that eventually led to artificial neurons that bear only superficial similarity to biological neurons and which employ implausible training methods, typically involving error backpropagation over large training sets. Nevertheless, for many important applications they work extremely well, and deep ANNs, in one form or another, have driven the machine learning juggernaut.

From the computer architecture perspective: given very complicated biological "hardware", the architect's job is to employ modeling and abstraction to separate those elements that support the basic computational paradigm from those elements that are there for other reasons, for example to provide a stable, reliable physical platform. The architect/researcher has wide discretion in divining the elements that are fundamental to the computing paradigm, and that makes the task both challenging and appealing.

*Second*: Once discovered, the paradigm can be implemented directly in silicon, using synchronous digital CMOS. As defined here, a *direct implementation* includes a silicon neuron for each model neuron, a synapse for each model synapse, and a wire for each connection in the model. The computing model is temporal, and communication is via unit time pulses (spikes).

The proposed direct implementation uses the CMOS synchronizing clock to mark the model's time units. This is a fundamental shift in traditional architecture abstraction layers. Ordinary, the hardware clock is at an abstraction layer beneath functional operation, so computed results are independent of the number of clock cycles required to compute them. In the architectures studied here, the hardware clock is at the same level of abstraction as functional operation: in effect, the number of clock cycles it takes to compute a value *is* the value.

As noted above, this paper primarily addresses the first hypothesis. However, the second hypothesis is an important consideration, and very low precision discrete mathematics ties the two together. Computation using very low precision (3 to 6 bits, say) is consistent with biological neural capabilities and is key to a feasible direct hardware implementation. Very low precision also reduces learning times dramatically, literally by orders of magnitude, because very low precision weights do not require lengthy and highly refined tuning. Finally, because the architect/designer is working with small, discrete values, the design process becomes combinatorial and somewhat *ad hoc* in nature. This makes the design process much closer to conventional digital design than it is to algorithm development based on the mathematics of reals and floating point software, as typically done in both neuroscience and machine learning. In fact, a computer architect/designer with only rudimentary knowledge of conventional machine learning should readily understand the proposed architecture.

### B. Near term applications

In the near term, online unsupervised clustering is a kernel function for many important edge-processing tasks. Examples include: 1) removing noise from sensor data, 2) compressing sensor information prior to sending it to a host process, thereby reducing transmission energy, 3) pre-processing data, thereby reducing the amount of AI processing performed at the host, 4) detecting anomalous behavior and triggering a supervisor process that intervention is required. A direct hardware implementation of these tasks could be implemented as a special purpose functional block in a system on a chip.

### C. Contributions

This research demonstrates both the inner-workings and potential of online TNN-based architectures, where the kernel function is online unsupervised clustering. To complete a basic system for performing online classification, a supervised TNN-like decoder maps cluster identifiers to output classes. This research:

1) Establishes a TNN architecture as a centroid-based clustering method. In effect, trained synaptic weights encode cluster centroids, and TNN inference computes the nearest centroid.

2) Demonstrates temporal computation enabled by ramp integrate-and-fire (RIF) neurons. Cluster membership is indicated by the presence of a spike, and distance to the cluster's centroid is indicated by the timing of the spike.

3) Demonstrates a simple learning method using spike timing dependent plasticity (STDP) that has the potential for greatly improved learning speeds (orders of magnitude better) when compared with more conventional methods that perform online supervised classification.

4) Proposes a new decoding method that uses a simple online supervised method to map TNN clusters into classes. Conceptually, it is a vote/tally method where voters learn through a supervised STDP-like mechanism.

## II. PRIOR RESEARCH

Spiking neural networks (SNNs) comprise two classes of networks that use spikes, or pulses, to encode and process information. One class, the TNNs considered here, encode information as precise timing relationships among individual spikes. This contrasts with the other class of SNNs that encode information as spike rates [13][28][30][32][43][48][49]. Rate-based encodings require roughly an order of magnitude more spikes to encode the same information, making them both slower and less energy efficient. Their slow processing speed alone provides a very strong argument against biological plausibility [54]. Some SNNs combine rate coding with STDP [4][10][17][18][44][52] or an autoencoder [45] to achieve unsupervised learning.

Of the TNNs, many train via offline backpropagation methods [9][15][39][46][59] that are typically transported or transformed from conventional convolutional neural networks (CNNs), making them both compute-intensive and unsuitable for online implementations.

In 1995, Hopfield [27] proposed a temporal coding method where training synaptic weights essentially tunes delays to match input spike patterns. Hopfield further posited that model neurons exhibit radial basis function (RBF) behavior, where output spike timing indicates how closely a given input pattern matches the associated pattern of trained weights. Neural networks exhibiting RBF behavior were proposed by Natschläger and Ruf in 1998 [42]. These networks have many features of TNNs considered here, albeit with significant differences in the details, including the temporal encoding method. The STDP method was relatively unstable, so a training method based on backpropagation was proposed and studied by Bohte et al. [8]. Despite this relatively early work, approaching the problem from the centroid-based clustering perspective has received little attention since, with the exception of the Tempotron [25] which uses supervised STDP to construct cluster centroids.

An important line of TNN research is due to Simon Thorpe and his academic descendants [7][26][29][37][41][53][57] as well as other researchers [19][58]. The research in this paper is most closely aligned with that body of work. The approaches described in [41] and [53] are singled out for comparison with the TNN systems proposed here (in Section VII.E).

Some SNN research is directed at platforms that support spiking neuron emulation but are relatively agnostic with respect to specific neuron models and methods. They have lots of designed-in flexibility so they can support research on a wide variety of models. Platforms range from simulators to special-purpose hardware. Most closely related to research in this paper is special-purpose hardware constructed with conventional CMOS. The IBM TrueNorth system is a prime example [3][14]. The Intel Loihi [16] is a more recent effort. The model proposed in this paper could be implemented on these platforms, with the amount of effort depending on the platform.

## III. SYSTEM ARCHITECTURE

TNNs are studied in the context of a system architecture that performs online supervised classification (see Figure 1). The *Supervised Learning (SL) System* continually observes a discrete time sequence of input patterns, $x(s)$, where $x$ is a vector and $s$ is the sequence number. The system processes each member of the input sequence in an online fashion and outputs a discrete time sequence $z(s)$ that indicates the *class* to which the input is predicted to belong. In an application environment, the "class" may invoke an external action, or it may provide a binary vector for further downstream processing. A sequence of supervisory inputs $I(s)$ is provided in response to the outputs. In this basic model, these indicate the correct class for each input.

In *online* systems as considered here, the learning algorithm operates sequentially as inputs arrive. For each input, the algorithm uses the supervisory inputs $I(s)$ for updating internal state (weights) and then moves on to the next sequential input. Inputs are not buffered for batch or mini-batch processing. A key feature of online systems is that they are readily amenable to processing realtime streaming input and can adapt dynamically as input patterns change at a macro level.

### A. Overview

The SL System includes five major subsystems as illustrated in Figure 1.

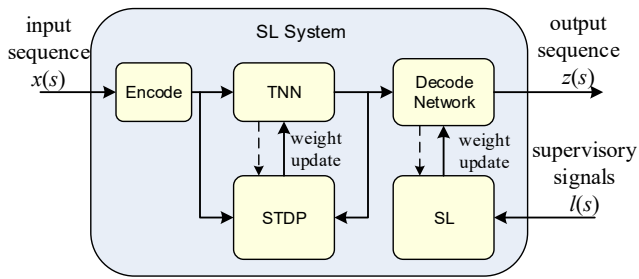

**Figure 1. Online Supervised Learning System. An unsupervised clustering front-end feeds a supervised decoding backend. Note there are no signals from the Decode Network back to the TNN. The dashed arrows indicate the reading of synaptic weights as part of the learning process.**

The *Encode Block* translates observed inputs from the environment into a temporal form for TNN processing. For our purposes, the system inputs may be conventional binary vectors transmitted over some type of bus. These binary vectors may originate at a sensory device such as a camera or microphone. Or they may be binary vectors coming from an upstream digital device. As such, encoding is highly dependent on the nature of the observed inputs. In many applications, encoding will be a hybrid process that converts purely spatial inputs into temporal signals sent to the TNN.

The *TNN* is composed of a hierarchy of *columns* that employ temporal processing to partition input patterns into clusters via online unsupervised learning. The term "column" is chosen because the scale and computational capabilities are roughly the same as a biological column in the neocortex [40]. However, it is not suggested that the internal organization and interconnections are the same. The learning method is a relatively simple form of STDP, which uses only locally available information to adjust weights as an ongoing process.

The unsupervised TNN feeds a supervised *Decode Network* that takes temporal spike volleys produced by the TNN's last-layer columns and maps them to a class. In a typical application, the class information is communicated to external devices or systems.

As the system's outputs are delivered to the external environment, supervisory signals are fed back into a supervised learning (*SL*) mechanism that is part of the Decode Network. In a basic implementation, these signals might indicate the accuracy of the output or the value of the correct output (e.g. a label). The SL mechanism operates in an online fashion and uses the supervisory signals to adjust weights that are part of the Decode Network. The supervised method incorporates an STDP-like learning mechanism: it uses local information, *except* the global supervisory signals are fanned-out to all the weight update units in the Decode Network.

Each of the three stages operates in an online manner leading to online operation of the overall system.

### B. Organization and Timing

The system organization consists of three cascaded stages corresponding to the Encode, TNN, and Decode functions. Refer to Figure 2. Of primary interest in this research is temporal processing, which is shown in the blocks with darker shading.

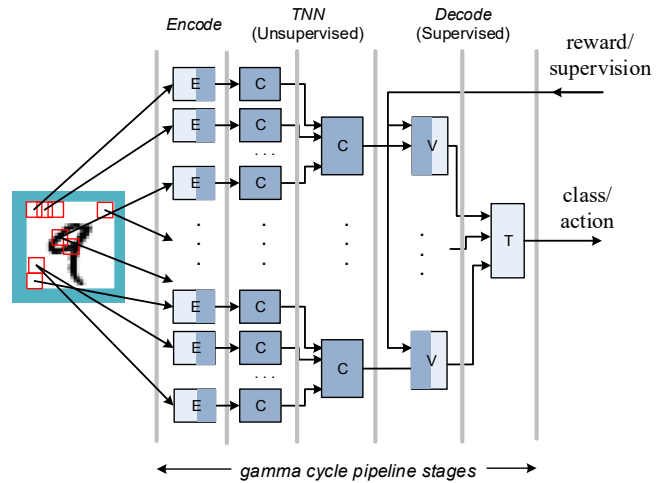

**Figure 2. Prototype architecture. E: encode, C: clustering column, V: supervised voter, T: tally block. Each communication path in the figure represents a bundle of lines in the implementation. In shorthand notation, this is an ECCVT network.**

Overall, the network acts as a *synchronous synfire chain* [1][2]. Parallel neurons are organized in layers and processing proceeds as *volleys* or waves of spikes pass from one layer to the next in a synchronized fashion. In this model there is at most one spike per line per volley. In the proposed architecture, the synfire chain is realized as a pipeline flow, synchronized according to a fixed-length version of the biological *gamma* cycle [21][24]. Thus, at a macro lev-el, each layer of the TNN is a single pipeline stage. Encode and Decode are likewise composed of one or more stages.

There are two clock levels in the architecture. The gamma pipeline cycle is at the macro level. At the micro level is the CMOS clock cycle which marks off model time units. In the implementation, a "spike" is a single cycle pulse that occurs sometime within the encompassing gamma cycle. The precise clock cycle in which it occurs encodes its value. The number of unit cycles per gamma cycle is defined to be $\tau_{max}$. In envisioned implementations, the value of $\tau_{max}$ is on the order of 8 to 16; i.e., temporal resolution is about 1-in-8 to 1-in-16. This very low precision is consistent with biological precision [12][36].

Temporal communication paths are *bundles* of *lines* (essentially parallel buses) that communicate spike volleys. A *volley* $x$ (bold) consists of $p$ spikes: $x = [x_1, \dots x_p], x_i \in N_0^\infty$. The set $N_0^\infty$ models discrete time units and consists of the non-negative integers plus the special symbol "$\infty$" which models the case where there is no spike during a given gamma cycle (conceptually, a spike that occurs infinitely far in the future). Figure 3 illustrates temporal communication and excitatory neuron processing. Each neuron operates according to its own local time, with computation beginning when it receives the first spike in a volley (local $t = 0$).

It is sometimes convenient to focus only on the presence of a spike, not its relative time. This is done by *binarizing* a volley. All lines with a spike ($x_i \neq \infty$) are converted to $x_i = 0$. E.g.: $binarize([0, 3, \infty, 1]) = [0, 0, \infty, 0]$.

## IV. NEURON MODELING

Excitatory neurons perform precise computation, and inhibition is modeled as a bulk, winner-take-all process.

### A. Excitatory Neuron Model

A spike response model (SRM)[22] is used in this work. Specifically, excitatory neurons employ *ramp-no-leak* (*rnl*) *response functions* $\rho(w,t)$ that map an integer weight $0 \leq w \leq w_{max}$ and an integer time $t$ onto the non-negative integers.

$$\begin{aligned} \rho(w,t) &= 0 && \text{if } t < 0 \\ &= t+1 && \text{if } 0 \leq t < w \\ &= w && \text{if } t \geq w \end{aligned}$$

Neuron $j$ generates output $y_j$ as a two-step process. First, spike-time-shifted response functions are summed at the body of neuron $j$ to yield a *body potential*, $\beta_j$:

$$\beta_j(t) = \Sigma \, \rho(w_{ij}, t - x_i) \text{ for } i = 1..p$$

Second, the output is produced via a *spiking function* $\sigma$ that compares the body potential against a threshold $\theta$:

$y_j = \sigma(\beta_j(t), \theta) = $ the smallest $t$ for which $\beta_j(t) \geq \theta$;
$y_j = \infty$ if there is no such $t$.

A typical biological response function has a rising edge that reaches a peak, then a falling edge that decays back to a baseline level. Computational neuron models differ in the way they handle the rising and falling edges. In some models, the response function rises as a single step, followed by a decay. Most leaky integrate-and-fire (LIF) models [50] [56] are defined this way. The step leading edge is convenient for event driven simulation and simplifies mathematical modeling.

Other models use simpler integrate-and-fire (IF) neurons based on *step-no-leak* (*snl*) response functions having a single unit time rise and no decay [19][20][29]. A supporting argument [41] is that the biological decay is primarily a reset mechanism. In simulation or hardware implementation, an explicit reset at the end of a gamma cycle can be done more simply than a decay throughout the cycle. The model used in this paper follows this method: there is no leak, and the gamma cycle initiates implementation state resets.

In contrast, the ramp integrate-and-fire (RIF) neurons used in this work are based on *rnl* response functions having a sloping leading edge. The ramp is key to achieving RBF behavior, to be expanded upon later.

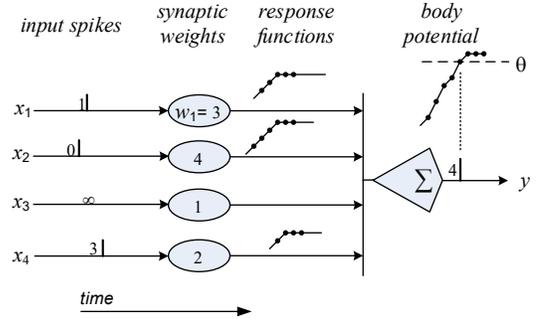

**Figure 3. Example of temporal encoding and excitatory neuron operation. Input values $x_1$ - $x_4$ are communicated as a *volley* of spikes conveyed over a *bundle* of four *lines*. Times increase from left to right, and all times are relative to the first spike observed by the excitatory neuron's summation function ($\Sigma$). The input volley encodes values [1, 0, $\infty$, 3], where $\infty$ is the assigned value when there is no spike on a given line. As the spikes arrive at the synapses, weights determine the amplitudes of response functions that are passed to the neuron body for summation; the response functions are shifted according to the relative spike times. At the neuron body, the time-shifted response functions are summed to form the body potential, and an output spike is generated the first time the body potential reaches the threshold $\theta$ (at $t = 4$). If it never reaches the threshold the output spike time is $\infty$.**

### B. Winner Take All Inhibition

Winner-Take-All (WTA) methods go far back in the annals of machine learning, and in one form or another WTA is ubiquitous amongst TNNs. Thorpe [55] first proposed that WTA inhibition should be performed on a bundle of lines carrying spikes. In the TNNs proposed here, one of the simpler forms of inhibition is implemented. An architected WTA inhibition block has the same number of input and output lines. Only the input with the earliest spike time is passed through uninhibited as an output spike. If there is a tie for the earliest spike, the tie is broken systematically by using the lowest index.

The inhibitory WTA vector function I has input lines $\boldsymbol{y}$ = $y_{1..q}$ and output lines $\boldsymbol{z}$ = $z_{1..q}$. Define $y_{\min} = min(y_i)$, $i = 1..q$. Then $\boldsymbol{z}$ = I($\boldsymbol{y}$), where

$z_i = y_i$  if $y_i = y_{\min}$ and for $k < i$, $y_k \neq y_{\min}$

$z_i = \infty$  otherwise.

## V. COLUMN ARCHITECTURE

The TNN is a layered, hierarchical network of interconnected *columns*. An example is in Figure 4.

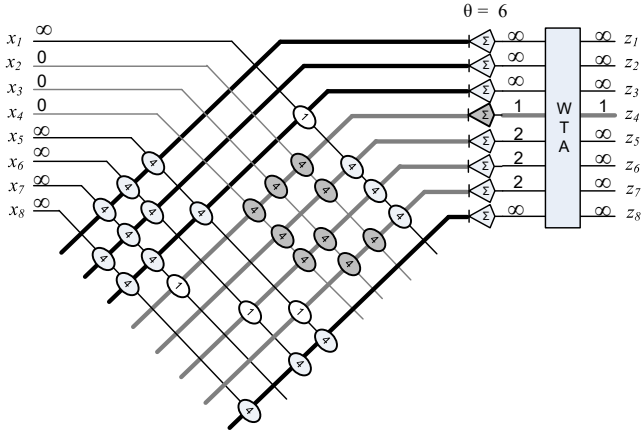

**Figure 4. A column consists of a synaptic crossbar feeding a column of parallel excitatory neurons. WTA inhibition selects the first output. The crossbar is drawn using an abbreviated notation: there is an individual line from each synapse to its associated neuron body. STDP adjusts weights according to input and output spike times. Synapses having θ weight are not shown. In this example, $w_{\max}$ = 4, and the synaptic weights are in a stable, bimodal distribution. The weights conceptually encode cluster centroids. For simplicity, binarized inputs are used in this example. RIF neurons are used. The fourth neuron (driving $z_4$) has three input synapses at $w_{\max}$ that receive spike inputs at $t = 0$. In this case, the body potential of $z_4$ to reach the threshold of 6 at $t = 1$. Three neurons have two synapses of weight $w_{\max}$ that receive input spikes, but the summation of their *rnl* response functions does not reach the threshold until $t = 2$. At the outputs, WTA inhibition selects the fastest spike, so only $z_4$ produces a non-∞ output value. Conceptually, this identifies the cluster centroid nearest to the applied input.**

Each column organizes input patterns into clusters based on similarity. There is one excitatory neuron per cluster. Conceptually, synaptic weights within a column encode cluster centroids. Informally, the *centroid* of a cluster is a dimension-wise average over each of the cluster members. The average is commonly the arithmetic mean, but other averages may be used. Synaptic weights are established by a learning process that is local to each column. The learning process does not literally compute the mathematical average; rather the result of learning is a statistical approximation to the average.

The output of each column is a bundle of lines $\boldsymbol{z}$, carrying a one-hot temporally encoded *cluster identifier* (CId) which indicates the cluster whose centroid is (approximately) nearest the applied input.

The layered hierarchy of columns forms clusters of clusters until the final layer of columns is reached. An important feature of the TNNs proposed here is that the final layer of the TNN may consist of *multiple* columns. Much (if not all) of the prior TNN research uses a single output column.

### A. Column Inference

Within a column, $p$ inputs are applied to $p \times q$ synaptic weight matrix $\boldsymbol{W}$, and $q$ excitatory neurons operating in parallel perform excitatory function E that produces a $q$ element output vector $\boldsymbol{y}$. I.e., $\boldsymbol{y}$ = E($\boldsymbol{x}, \boldsymbol{W}$). It is assumed that during operation the threshold θ is fixed and therefore is not shown as a function input.

The inference process determines cluster membership by evaluating the set of vector functions $\boldsymbol{z}$ = I(E($\boldsymbol{x}, \boldsymbol{W}$)) which map inputs to a set of $q$ clusters, $C_{1..q}$ as follows: input pattern $\boldsymbol{x} \in C_i$ if there is a spike on line $z_i$, i.e., $z_i \neq \infty$. Note that it is not required that every pattern $\boldsymbol{x}$ belongs to a cluster; i.e., for some patterns, $z_i = \infty$ for all $i$.

### B. STDP training

The STDP strategy strives for weights that stabilize at a bimodal distribution, where the predominant weights are 0 and $w_{\max}$. The functional objective of STDP is to partition a stream of input patterns into clusters according to similarity and associate each cluster with an excitatory neuron. The values of a neuron's synaptic weights are highly correlated with the neuron's cluster's centroid.

STDP is implemented as a parallel set of small finite state machines, one at each synapse. As each input pattern is applied, the synaptic weight $w_{ij}$ is updated for the next sequential input using only the synapse's input spike time $x_i$, the output spike time $y_j$ of its associated neuron, and the current value of $w_{ij}$. The STDP update function either increments the weight by $\Delta w_{ij}$ (up to a maximum of $w_{\max}$), decrements the weight by $\Delta w_{ij}$ (down to a minimum of 0), or leaves the weight unchanged. Table 1 defines the STDP update function for a given synaptic weight $w_{ij}$.

**Table 1. STDP Update Function**

| input conditions | | weight update |
|---|---|---|
| $x_i \neq \infty$ | $x_i \leq y_j$ | $\Delta w_{ij} = + \mu_+ \cdot F_+(w_{ij})$ |
| $y_j \neq \infty$ | $x_i > y_j$ | $\Delta w_{ij} = - \mu_- \cdot F_-(w_{ij})$ |
| $x_i \neq \infty$ | $y_j = \infty$ | $\Delta w_{ij} = +\mu_s$ |
| $x_i = \infty$ | $y_j \neq \infty$ | $\Delta w_{ij} = - \mu_- \cdot F_-(w_{ij})$ |
| $x_i = \infty$ | $y_j = \infty$ | $\Delta w_{ij} = 0$ |

STDP input conditions are divided into four major cases, corresponding to the four combinations of input and output spikes being present ($\neq \infty$) or absent ($= \infty$). When spikes are present on both, two sub-cases are based on the relative timing of the input and output spikes in the classical STDP manner [6][33][38][23].

Classically, STDP update is a function of both $\Delta t = x_i - y_i$ and $w_{ij}$ [6][38]. In the model proposed here, the temporal component is a constant: $\mu_+$, $\mu_-$, or $\mu_s$. The functions of the current weight, $F_+$ and $F_-$, are defined as:

$F_+ (w_{ij}) = \mu_+$   if $w_{ij} \geq w_{max}/2$; else   $F_+ (w_{ij}) = \mu_+/2$

$F_- (w_{ij}) = \mu_-$   if $w_{ij} < w_{max}/2$; else   $F_- (w_{ij}) = \mu_-/2$

The characteristic of $F_+$ is that if the current weight is in the upper half of its range, then the full increment $\mu_+$ is applied; if it is in the lower half of its range only a half increment is applied. Thus, a weight in the upper half of the range is biased to stay there. This provides some stability and tends to speed up the weight convergence process. $F_-$ is biased the opposite way.

Aside: the *search* mode is new in this work although it doesn't really come into play for the benchmark results given here. There is an increment of $\mu_s$ if a synapse receives an input spike but its neuron generates no output spike. In the absence of a search mode, a weight is only incremented if its associated neuron emits an output spike ($y_j \neq \infty$). Consequently it may be possible to reach a state where a neuron's synaptic weights are all relatively low, so the neuron never yields an output spike for any input pattern. The weights are permanently stuck at their low values. With search mode, $\mu_s$ is typically a small value, and if input patterns are applied repeatedly without an output spike, eventually weights will creep up to a level where an output spike is produced.

## VI. DECODE

The Decode Block consists of a layer of *voters* followed by a *tally* block. This decode method is new in this work.

### A. Voter

*Voters* are associated 1-to-1 with TNN output columns. Refer to Figure 5. A voter takes the column's output $z$ (encoding a CId), and outputs a bundle $\boldsymbol{v}$ carrying one or more binarized spikes indicating a set of classes that are more frequently associated with that cluster's input patterns. These are considered as "votes" for the correct class, and a single column can cast multiple votes. The implementation generates votes by constructing what amounts to a lookup table, implemented as a weight matrix, with the CId as the table index. In the hardware, the weights are learned with online supervision in an STDP-like manner.

The last layer columns produce *temporal one-hot* CIds, $z_{1..q}$. The CIds are one-hot in the sense that there is only one spike and are temporal in the sense that the spike time may encode a range of values. Recall that the gamma cycle consists of $\tau_{max}$ unit cycles. If there are $q$ output lines and their temporal range is 0 to $\tau_{max} -1$, then there are $q*\tau_{max}$ different temporal CIds. To reduce implementation cost with little effect on accuracy, the voter reduces time $\tau_{max}$ to a smaller effective time window $\tau_{eff}$. That is, if output spike time $y_i > \tau_{eff}$ then $y_i$ is set equal to $\tau_{eff}$. The early output spike times are unchanged; all the later spike times are mapped to $\tau_{eff}$.

If there are $r$ classes, the output signals $v_{1..r}$ encode votes for the "correct" class, and the number of votes cast by a given voter can range from 0 to $r$, depending on the probability that the CId is associated with members of a given class.

Voter training is supervised via the signals $\boldsymbol{l} = l_{1..r}$ : a binary one-hot encoding of $r$ classes. If $l_i = 0$ (a spike at $t = 0$), then class $i$ was correct for the immediately preceding input. This information is used for updating the weights held in the voter in a localized STDP-like manner.

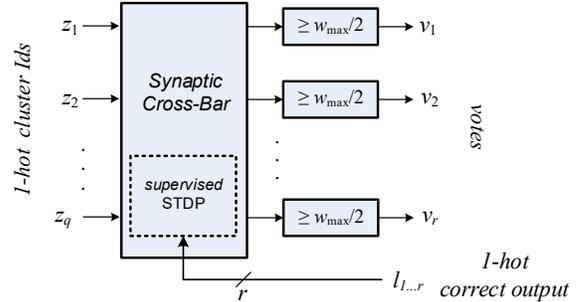

**Figure 5. A voter maps 1-hot CIds to votes. The supervised STDP method takes $z$ and $l$ as inputs, but not $v$. Because inputs are 1-hot, no weight summation is required: a single weight is read and compared with the fixed threshold $w_{max}/2$.**

Ideally, a voter performs the following function. Given a cluster identifier where $z_j = k$, a vote is cast ($v_i = 0$) for any and all of the $r$ classes whose probability of being correct reaches a fixed threshold $0 \leq \theta_v \leq 1$. More formally, if $P(l_i = 0 \mid z_j = k) \geq \theta_v$ then $v_i = 0$; else $v_i = \infty$.

Of course, the implementation can't divine the true probability. Nevertheless, one can use *ad hoc* means to engineer a logic circuit that provides a good approximation. This is done in the next subsection.

### B. Implementation

The voter implementation employs a $q \times r$ crossbar with $\tau_{max}$ low precision saturating up/down counters at each of the crosspoints: a total of $q \times r \times \tau_{eff}$ counters. The crossbar inputs are the $z_i$, and the crossbar outputs are the $v_j$.

The counters hold weights that range from 0 to $w_{max}$. Table 2 specifies the weight update functions for the $k$ counter at crosspoint $i, j$.

**Table 2. Voter Weight Update Function**

| input conditions | weight update |
|---|---|
| $z_i = k$; $l_j = 0$ | $\Delta w_{lijk} = +(1 - \theta_v)$ |
| $z_i = k$; $l_j = \infty$ | $\Delta w_{lijk} = -\theta_v$ |
| $z_i = \infty$ | *no update* |

With this method, if the probability of label $l_j$ is greater than the threshold value $\theta_v$, then the counter will tend toward $w_{max}$ and saturate there. If less, it will tend toward 0 and saturate there.

Observe that update depends only on the CId received from the TNN and the supervisory signals (e.g., the correct label); update is independent of any votes that were cast.

### C. Voter Inference

Given a trained synaptic network, the voting process uses the one-hot input CId, $z$, to access crosspoints connected to the outputs $v$. If the spike time on line $z_i$ is $k$, the value of counter $k$ at crosspoint $i$, $j$ is compared with $w_{max}/2$ to produce a vote (a spike on $v_j$ at $t = 0$).

$v_j = 0$ iff $(z_i \equiv k) * w_{ij} \geq w_{max}/2$.

### D. Tally Block

The *tally* block sums all the votes from multiple voters. It yields a single "winner" unless there is a tie, in which case there may be multiple winners -- or a tie-breaker. The tally block operates by rote: there is no training involved, it simply adds up votes and declares the winner. At the gate level, the tally block can be implemented in a straightforward way as a parallel counter [51].

### E. Multiple Thresholds

By defining the voting threshold $\theta_v$ to be a relatively high value, then only classes that have a high probability of being correct receive votes. Hence, one strategy is to set a high $\theta_v$ and tally only high likelihood votes. The intuition is that combining a number high likelihood votes will greatly increase the overall likelihood of choosing the correct winner.

Another, less obvious strategy is to set $\theta_v$ to be very low, so that only very unlikely classes do not receive a vote. Then, when votes are tallied, the class that receives the fewest "non-votes" will be the winner. In effect, the winner is decided by a process of exclusion.

It has been found through experimentation (Section VII.F) that the low threshold method is the better of the two (by a wide margin), but a combination of the two works better still. Consequently, two voters are implemented: one with a high threshold $\theta_{Hi}$ and one with a low threshold $\theta_{Lo}$, then both sets of votes are tallied together to determine the final "winner".

## VII. Prototype System

This section describes a prototype system and demonstrates its operation.

### A. Proof-of-Concept Benchmarks

The proposed architecture is demonstrated using two benchmarks, both based on the MNIST dataset [31]. Realistically, MNIST is not an application that requires online learning. Nevertheless, experimenting with the MNIST dataset has significant advantages: the MNIST dataset is easily explained, very well studied and understood, and it has a level of complexity that is challenging, but not overly so. Because MNIST inputs are independent (it is assumed), they are adequate for researching feedforward networks.

Although the dataset is used, the MNIST benchmark protocol, which conforms to an offline learning scenario, is not.

For an online scenario, the 60K train set and the 10K test set are merged into a 70K stream that is applied once to form a 1-*phase* benchmark. A 3-*phase* benchmark sequence is designed to demonstrate dynamic adaptability by modifying the original 70K stream and dividing it into three phases:

1) The first 20K MNIST images are applied as-is.

2) The second 20K images are transposed -- flipped along the diagonal. The labels are unchanged. This phase demonstrates the ability of the unsupervised TNN to adapt to radically changed input patterns.

3) The final 30K images are both transposed and have odd and even labels switched. I.e., all 0s are re-labeled 1's, 2's re-labeled 3's, etc. This phase tests the adaptability of online supervised learning implemented in the voter.

Regarding metrics, *error rate* is typically used as a classification metric. Because the focus is on online dynamic operation, error rate alone is not very revealing because the error rate is a function of the input sequence length. Hence, in this work, error rates are accompanied by the input length required to achieve that rate.

### B. Encode

At this stage of the research, TNN temporal processing is of utmost interest, and this influences the choice of the prototype encoder. An encoder pre-processes input data before passing it to the TNN, and much of this pre-processing may be non-temporal. Consequently, for the research prototype, the encoding function is kept very simple in order to place a greater burden on TNN processing and less on encoder preprocessing. This avoids a system where lots of computation is done in encoding and relatively little in the TNN. The online decoding method is simple for similar reasons.

To keep the encoding process simple, the MNIST patterns are first binarized by removing grayscale. Then, the encoder generates spikes for both positive and negative versions of a given input pattern (Figure 6). Consequently, the number of spikes per encoded pattern is constant. This yields a form of homeostasis, and one-hot coding keeps spike counts more-or-less in balance throughout the network, although much sparser.

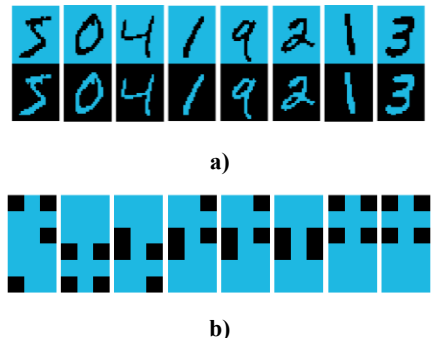

**Figure 6. First 8 MNIST images, PosNeg encoded and binarized. A black pixel indicates a spike at *t*=0; a blue pixel indicates no spike. a) original 28×28 image, b) Layer 1 3×3 receptive fields at location [14,14] with corner pixels PosNeg encoded to form a 6×3 input image.**

Only the TNN's primary inputs coming from the encoder are binarized. Internal to the TNN, the column outputs are in temporal form (Figure 4). Hence, by studying column outputs, the features of TNN temporal processing are more clearly revealed (Sections VII.E and F).

At the input, an encoder selects the four corner pixels of 3×3 receptive fields (RFs). After forming the negative, there 8 pixels in all (Figure 6b), exactly 4 of which translate to spikes at $t = 0$. This approach was determined to work well, as it provides a modicum of spatial invariance.

### C. System Configurations

Three system are studied: an ECVT system comprised of an **E**ncoder, a single layer of **C**olumns, a layer of **V**oters and a **T**ally network, and ECCVT and ECCCVT systems with two and three layers of columns respectively. The first C layer contains a column for each of the 676 overlapping 3×3 RFs taken from the binarized 28×28 MNIST images. There are 12 neurons per column (hence 12 clusters). Layers 2 and 3 are similarly constructed with details given in Table 4. The voter layer consists of either one or two voters (either a low threshold voter or both high and low thresholds). Synapse counts for the three systems are given in Table 3. Totals are given only for 2 voter systems.

**Table 3. Experimental System Configurations**

|  | inputs | outputs | number | $\tau_{eff}$ | total syns |
|---|---|---|---|---|---|
| Layer 1 | 8 | 12 | 676 |  | 65K |
| Layer 2 | 12 | 20 | 576 |  | 553K |
| Layer 3 | 20 | 32 | 484 |  | 1239K |
| L1 Voter (each) | 12 | 10 | 676 | 2 | 162K |
| L2 Voter (each) | 20 | 10 | 576 | 3 | 346K |
| L3 Voter (each) | 32 | 10 | 484 | 4 | 620K |
| **ECVT total = 65K + 2*162K** |  |  |  |  | **389K** |
| **ECCVT total = 65K + 553K + 2*346K** |  |  |  |  | **1310K** |
| **ECCCVT total = 65K + 553K + 1239K + 2*620K** |  |  |  |  | **3096K** |

**Table 4. Column and Voter Simulation Parameters**

|  | column parameters | | | | voter parameters | |
|---|---|---|---|---|---|---|
|  | $\theta$ | $\mu_+$ | $\mu_-$ | $\mu_s$ | $\theta_{Hi}$ | $\theta_{Lo}$ |
| Layer 1 | 4 | 1/2 | 1/2 | 1/1024 | 15/32 | 1/32 |
| Layer 2 | 8 | 1/4 | 1/4 | 1/512 | 21/32 | 1/64 |
| Layer 3 | 8 | 1/4 | 1/4 | 1/512 | 24/32 | 1/64 |

Column and voter parameters are in Table 4. Non-integer parameters are given as fractions. As noted earlier, the weights are integers that range from 0 to $w_{max}$. $\mu$ values less than 1 imply there is a lower-order counter fraction that does not affect $\tau_{max}$ or the number of response functions. Hence, if the parameter is 1/4, this suggests 2 additional counter bits to the right of a fixed point. As noted earlier, as the parameter $\mu_s$ doesn't come into play with this benchmark and is kept at a very low background level. All weights are initialized at $w_{max}/2$.

### D. Primary Results

The 1-*phase* ad 3-*phase* benchmarks were streamed through the network, and error rates at 1K intervals were measured. Learning is concurrent with measurement. Figure 7 plots results for the three studied systems. Note that because the research emphasis was on function, no special effort was made to reduce learning times.

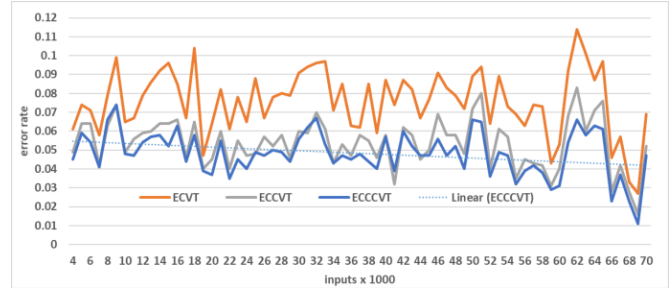

a) 1-phase benchmark 4K-70K inputs.

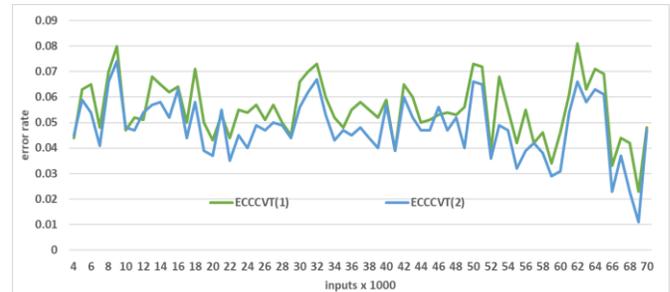

b) ECCCVT system with 1 voter and 2 voters.

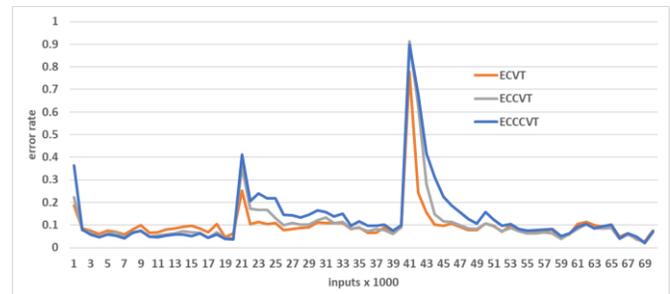

c) 3-phase benchmark

**Figure 7. Benchmark results for prototype system. Unless noted otherwise, systems use 2 voters.**

Figure 7a plots error rates for the 1-*phase* input stream. This plot begins at 4K inputs to permit a finer granularity *y*-axis. (The error rates for the first 3K are in Figure 7c). All three systems learn quickly, reaching an error rate well under .1 by the time 4K inputs are reached. A linear trend line for the ECCCVT system shows error rates heading downward as more inputs are applied. The interval-to-interval variations are a natural effect of fast online learning; infer-ence is based only on the recent past so synaptic weights are constantly in flux.

Figure 7b plots error rates for the single voter system (ECCCVT(1)) and the two voter system (ECCCVT(2)). Accuracy mostly depends on the low threshold voter (i.e., the absence of low threshold votes). The high threshold voter contributes incrementally to accuracy. A potential way of reducing cost is to use a low threshold voter only.

Figure 7c plots results for the 3-*phase* benchmark. The times when the images are transposed (at 20K) and labels are switched (at 40K) are very evident, as is an expected learning time lag between the three C layers. In all cases, the network learns the new patterns/labels quickly, although the disruption is bigger when labels are switched. For transposed images, it takes between 3K and 10K instructions to re-stabilize, depending on the depth of the network. For the label switches, it is 4K to 12K.

### E. Comparisons

Comparison with recently published results are in Figure 8. Selected points are shown for the ECCCVT system.

Conventional offline machine learning methods based on backpropagation have been adapted to work in an online fashion. Some version of online gradient descent (OGD) [60] is commonly used. Sahoo et al. [47] propose a new method, Hedge Backpropagation (HBP), and they evaluate a variety of other OGD-based systems for comparison. One set of evaluations is for a synthetically extended version of MNIST [34]. The best of the methods (better than the baseline OGD methods) is a 20 layer deep neural network using HBP.

In Figure 8 three data points are taken from Figure 2c in [47] for the HBP method. The chart in [47] tops out at an error rate of .09, and the HBP method requires 100K inputs to get beneath that level. An error rate of .04 is reached after 500K inputs. Hence, to reach error rates achieved by the ECCCVT system requires a training sequence 100 times longer. Finally, after 5M inputs, the HBP method achieves an error rate of about .015: the result of very lengthy, high precision weight training.

In his MS thesis, Büller [10] describes an OGD method that uses rate-based spiking neurons. He states this method is closely related to a method proposed by Bellec et al. [4] and gives results for both methods. The training set is run for 30K inputs before accuracy is measured. For the data points provided by Büller, this method is less accurate than the ECCCVT system, even after an order of magnitude more training inputs.

Of the prior TNN implementations, the two most similar to the one proposed here are presented in Mozafari et al. [41] and Thiele et al. [53]. The system in [41] consists of an encoder that uses 6 difference of Gaussian (DoG) filters, and the system in [53] uses spike rates to encode grayscale. Both use STDP methods similar to the one proposed here. Both use the simpler IF neuron model, unlike the RIF model used here. Both use TNNs with two convolutional layers employing a machine learning-inspired feature map organization followed by global STDP-like decode layer. In contrast, TNNs proposed here employs a column organization: each first layer column learns solely based on its own RF. Both systems [41] and [53] use pooling layers (although different from each other); there are no pooling layers in the TNNs proposed here.

Both [41] and [53] include features that appear biologically implausible. The system [41] incorporates weight sharing over synapses that would be physically separated in the biology. The method in [53] uses neuron bodies with two integrators (and two thresholds) per neuron: one for STDP and the other for inference.

Although not explicitly constructed or evaluated as an online system, the network in [41] can be adapted for online operation in a straightforward way. The first layer is trained with 100K patterns, the second layer with 200K, and the supervised layer with 40M. The MNIST error rate after 40M training inputs is .027. This is slightly better than the error rates achieved with the ECCCVT system, with 2+ orders of magnitude more training inputs. The authors of [41] appear to have been striving for their best achievable accuracies, most likely at the expense of very long training times.

The system in [53] is explicitly designed for online operation, and is the closest in overall performance to the method proposed here (see [53] Fig. 6a), although after the same number of inputs, the error rates are 25-50% higher .

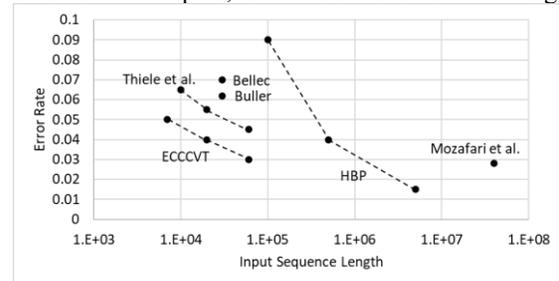

**Figure 8. Learning is about two orders of magnitude faster than other online approaches having similar error rates.**

### F. Secondary Results

#### RIF neurons

The ECCCVT system was simulated with IF neurons in place of RIF neurons. Results are in Figure 9. Error rates generally track each other. However, with RIF the error rates are less than half the error rates with IF.

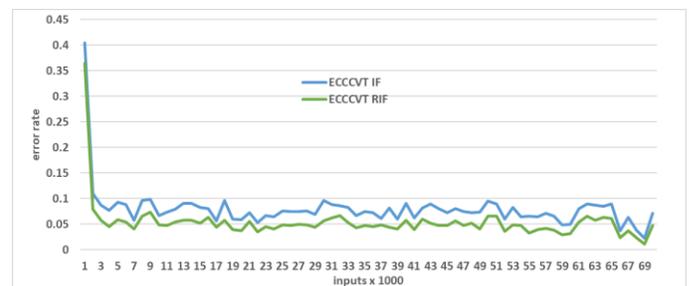

**Figure 9. Error rate of IF and RIF neurons.**

If synaptic weights have the same ranges, gate level implementations of IF and RIF neurons will have very similar gate counts because values of the same magnitude are summed.

A classic theoretical result due to Maass [35], is that a step response is functionally less capable than a ramp response (subject to a set of reasonable assumptions). This theoretical result is aligned with the experimental results just given.

Finally, observe that if an IF neuron is used in conjunction with binarized input patterns (as done in this paper), then the network degenerates to the classical binary perceptron. Consequently, the binary perceptron can be viewed as a common ancestor of both mainstream ANNs and TNNs.

### Weights encode cluster centroids

To establish that the proposed model is essentially a centroid-based clustering method, a simple metric for centroid convergence is used. After a column has mapped input patterns into clusters, the *centroid* $c^k$ of cluster $C_k$ is an element-wise arithmetic mean of the members of $C_k$.

After the centroids of the column-mapped clusters are established, the centroid nearest to a given input pattern is computed. In this paper, the distance metric is sum of absolute differences: $sad(x,y) = \Sigma_{i=1..p} \mid x_i - y_i \mid$; this is a rectilinear (Manhattan) distance metric[1].

Define centroid convergence metric *c_conv* as the fraction of input patterns whose computed nearest centroid is the one associated with their column-mapped cluster. This metric is drawn from the convergence criterion for classic *k*-means clustering, except here it is not used for testing convergence as the method proceeds. Rather, it is only applied after clustering is finished to measure the quality of clustering that has been done. The value of *c_conv* = ranges from 0 to 1, with 1 indicating ideal convergence.

To evaluate this metric, clusters were determined for the input patterns between 60K and 70K of the 1-*phase* MNIST dataset. Their centroids were computed, and *c_conv* was computed by scanning through input the patterns. For network layer 1, *c_conv* = .999, for layer 2, *c_conv* = .981, and for layer 3, *c_conv* = 962. These numbers strongly support the assertion that STDP uses synaptic weights to encode approximate centroids and that the inference process computes an approximation to the nearest centroid.

### RBF Behavior

To demonstrate RBF behavior, three RFs from the ECCVT network were simulated. Based on 28×28 images with [1,1] in the upper left corner, three 5x5 RFs at [6, 12], [12, 12], and [18,12] were simulated using the 1-*phase* benchmark. After 60K initial inputs, the subsequent 10K outputs were analyzed with results plotted according to output spike time (Figure 10).

Output spike times range from 0 to 7, although no spikes occur at $t = 4$ or 5, and a negligible fraction occur at $t = 7$ (Figure 10b). An earlier spike time is highly correlated with a lower distance (measured as *sad*) from the centroid. This is the claimed RBF behavior.

If one were to use a IF response function for this same case (recall that all inputs to layer 1 are binarized) all inputs would occur at $t = 0$, and all output spikes would appear at $t = 0$. There would be no RBF behavior. Or in other words, RBF behavior results from temporal processing brought about by the RIF neuron.

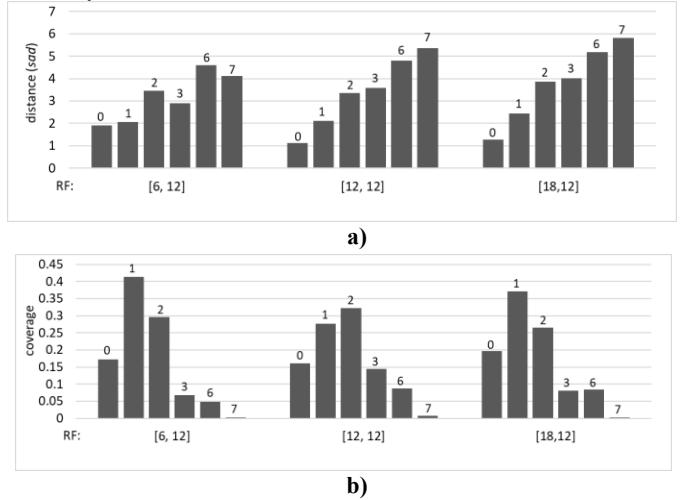

**Figure 10. Plots illustrating RBF behavior for three selected RFs of ECCVT. Output spike times label the tops of the bars. a) The earlier the spike time, the closer the input is to the centroid, measured as average *sad*. This is RBF behavior. b) Coverage is the fraction of input patterns that result in the given output spike time. Spikes at *t*=7 are rare, with spikes at *t*=0, 1, and 2 dominating.**

## VIII. DISCUSSION AND FUTURE RESEARCH

### A. Support for the Two Hypotheses

Consider the two hypotheses stated in the introduction.

Significant support is provided for the *first* hypothesis. The main components of the model all come from the neuroscience literature and are supported by strong plausibility arguments: synfire chains [1], temporal coding [27], SRM excitatory neurons [22], WTA inhibition [55], column architecture with RBF functionality [42], and STDP [23][33]. The proof-of-concept simulations demonstrate that a TNN based on these primary components is capable of efficient, dynamic, flexible operation.

Regarding the *second* hypothesis on the feasibility of a direct CMOS implementation, first observe that all the TNN components involve straightforward logic design. Regarding feasibility, some order-of-magnitude estimates can be advanced.

Virtually all the cost is in the synapses (there are 100 times as many synapses as anything else). Say each neuron has $10^2$ synapses, implemented as low precision saturating up/down counters. Conservatively, each consumes $5 \times 10^2$ transistors, so each neuron accounts for $5 \times 10^4$ transistors.

---

[1] Using Euclidian distance changes results insignificantly.

Today, CMOS can be fabricated with 100M transistors per mm$^2$, and a chip may have an area of 500 mm$^2$. Hence, $5 \times 10^{10}$ transistors per chip or $10^6$ model neurons. For comparison, the biological neocortex contains $10^5$ neurons per mm$^2$ and $25 \times 10^9$ total neurons. Consequently, it would take 25K silicon chips to match the biological neuron count. With today's silicon technology, this is not an infeasible number, but it is pretty large.

Considering speed, the biological gamma cycle is about 100 Hz (at the high end). A silicon gamma cycle is conservatively estimated to be 10 MHz -- a factor of $10^5$ faster. If a linear tradeoff between area (neuron count) and speed (gamma frequency) is assumed, then area × frequency is:.

$[10^6] * [10^7]$ (silicon) $vs \cdot [25 \times 10^9] * [10^2]$ (biology)

$\rightarrow 10^{13}$ (silicon) $vs\ 2.5 \times 10^{12}$ (biology).

By this simple area times frequency metric, there is potentially 4× more computing power on a single silicon chip than in the neocortex.

Finally, with regard to the MNIST application, percentages of modern-day chip area would be: ECVT-- .4 % ; ECCVT-- 1.3%; ECCCVT-- 3.1% . The MNIST "frame rates" would be $10^7$ per second.

*Caveat 1:* A smooth, linear area×speed tradeoff will be very difficult to achieve with a direct implementation because the cost of a context switch will likely be prohibitively high. Hence, a direct hardware implementation will likely be dedicated to a single task for extended periods of time. For some applications, advantage may be gained by trading area for power by using CMOS with very low static power consumption and running it at a very slow clock cycle. Furthermore, signaling is sparse. This may be a good approach for edge processing tasks that handle sensory input and have very strict power budgets.

*Caveat 2:* The interconnection network is ignored in the above analysis. To allow a direct implementation as envisioned here, the network will necessarily be circuit switched with controlled path lengths. So, in practice, the network is first initialized for a dedicated application by setting the circuit switches. Then operating neurons pass spikes deterministically and uninterrupted through the switches. Such a circuit switched network will be a design challenge and could easily consume as much area as the rest of the TNN.

*Caveat 3:* Energy consumption must be controlled. Every unit delay in the model implies a D flipflop somewhere in the CMOS implementation. There is no counterpart to these flipflops in the biological implementation. Synaptic state, in the form of saturating counters, dominates the hardware, and all this state is clocked. If left unchecked, the amount of energy burned in clock signals will be extremely high. Fortunately, clock gating opportunities appear to be widespread and systematic, offering hope that clock power can be constrained successfully.

*Caveat 4 (in the other direction):* Biological neurons and synapses are unreliable. Consequently, there is significant redundancy in the neocortex to compensate for the unrelia-bility. With CMOS, reliability is many orders of magnitude better; consequently the CMOS implementation may save multiple factors of neurons and synapses because high levels of component-level redundancy are not needed.

*B. Future Research*

Caveats 2 and 3 in the preceding section point immediately to areas for future research. Others follow.

***Improved, deeper non-recurrent networks.*** The voters and the way they are applied deserve further attention. First, although the voters are very simple -- nothing more than synaptic crossbars -- they encompass a significant fraction of total synapses. One way of reducing voter synapse counts is suggested above: use, and improve upon, a single threshold voter network.

Furthermore, the vote/tally process discards spatial information: a vote is a vote regardless of where it comes from. In the systems studied here, the voter/tally network spans a large column space, so a lot of spatial information is discarded. The column process, in contrast to vote/tally, is very good at recognizing spatial patterns. Consequently, an approach that shows some promise is to intersperse regional vote/tallies between column layers, thereby reducing spatial information incrementally. This has the look of a pooling function that supports spatial invariance. As a simple example, one might implement an ECVTCVT system.

***Reinforcement learning (RL).*** Applications based on reinforcement learning are prime candidates for approach described here. Supervised classification was used in the prototype to keep things simple. Moving to an RL version for non-recurrent networks is a (possibly short) next step.

***Recurrent networks.*** Ultimately, recurrent networks are the target and the place where TNNs may show their greatest strength. Feedback connections are implementable in a straightforward way: structured according to gamma cycles. With the STDP learning process, there is no need for back propagation through time; learning rules in the model respect the flow of time.

***Applications.*** As mentioned in the introduction, edge processing tasks seem to be prime candidates for initial application of directly-implemented TNNs. Longer term, recurrent systems are the real target, so development of simple applications that require recurrent systems is a prime research direction. For example, a good research benchmark is a predator/prey video game.

***CMOS chip design.*** If the first applications involve edge processing tasks, then initial proof-of-concept CMOS designs should be targeted at such tasks. Area, time, and power estimates will shed better light on the feasibility of direct implementations.

***Emerging technologies.*** Conventional CMOS is considered here, and synapse costs dominate. New technologies such as memristors and RRAM show some promise for much cheaper synapse implementations. However, a key question is whether they can efficiently implement the specific synaptic functions as proposed here.